\documentclass[letterpaper, 10 pt, conference]{ieeeconf}  

\usepackage{graphicx}
\usepackage{booktabs}
\usepackage{ amssymb }
\usepackage{amsmath}
\usepackage[ruled,vlined]{algorithm2e}
\usepackage{ mathrsfs }
\usepackage{mathtools}
\usepackage{flushend}
\usepackage{hyperref}
\usepackage{xcolor}
\usepackage{multicol}
\usepackage{subcaption}
\usepackage{siunitx}
\usepackage{float}
\usepackage{algorithmicx}
\IEEEoverridecommandlockouts                              

\title{\LARGE \bf
Maintaining the Level of a Payload carried by Multi-Robot System on Irregular Surface
}

\author{Rishabh Dev Yadav, Shrey Agrawal, Kamalakar Karlapalem
\thanks{Accepted at \href{https://sites.google.com/umd.edu/icra-2020-ws-multi-robot/program}{ICRA 2020 Workshop: Multi-Robot} ; 4 June, 2020. }
\thanks{Rishabh is a Research Assistant at Agents and Applied Robotics Group (AARG) at International Institute of Information Technology, Hyderabad (IIIT-H), India.
       {\tt\small rishabhdevyadav97@gmail.com}}%
\thanks{Shrey is a MS by Research student of AARG at IIIT-H.
       {\tt\small shrey.agrawal@iiit.ac.in}}%
\thanks{Kamalakar Karlapalem is a faculty of Computer Science with IIIT-H.
       {\tt\small kamal@.iiit.ac.in}}%
}

\begin{document}

\maketitle
\thispagestyle{empty}
\pagestyle{empty}

\begin{abstract}
In this paper, we introduce a multi robot payload transport system to carry payloads through an environment of unknown and uneven inclinations while maintaining the desired orientation of the payload.
For this task, we used custom built robots with a linear actuator (pistons mounted on top of each robot.
The system continuously monitors the payload's orientation and computes the required piston height of each robot to maintain the desired orientation of the payload.
In this work, we propose an open loop controller coupled with a closed loop PID controller to achieve the goal. 
As our modelling makes no assumptions on the type of terrain, the system can work on any unknown and uneven terrains and inclinations.
We showcase the efficacy of our proposed controller by testing it on various simulated environments \footnote{Simulation (using Gazebo) video can be found \href{https://youtu.be/mw2Bc3rAbrg?si=2b0SkTwLHn210b_D}{\textcolor{blue}{here}}.
} with varied and complex terrains.

\end{abstract}

\section{INTRODUCTION}
The problem of cooperative payload transportation has gained a lot of traction in recent years \cite{ohashi2016} \cite{bou2003} \cite{mars2002}. Increased attention and extensive work has been done on formation control \cite{c1}, \cite{c2} \cite{c3}. Applications of multi-robot payload transport systems range from indoor warehouse logistics \cite{kiva} to outdoor disaster search and rescue environments. Some scenarios require transporting large payloads quickly, safely, and inexpensively across complex uneven terrain \cite{baou2005}.
In multi-robot payload carrying system, we can assign a number of robots to carry a payload.
This makes the system modular so that we can control the number of robots in a team (formation) depending on the weight of the payload.
Thus, we can optimize the energy spent in transportation by assigning only an optimal number of robots to carry the payload.
A multi-robot system is also capable of generating high traction torque which is advantageous while moving on an inclined plane.
This kind of system can also be made failure resilient under some controlled conditions as shown in \cite{pulkit}.
In \cite{bou2003} and \cite{baou2005}, only 1-D payloads and two mobile robots are used, which leaves out a lot of scope of multi robot payload transport.
\cite{mars2002} also presented a similar concept with decentralized coordination. Any of \cite{bou2003}, \cite{baou2005} and \cite{mars2002} doesn't focus on 2-D payload and maintaining it's orientation.
In our work, we explore the problem of carrying a payload by a multi-robot team on uneven terrains and inclined surfaces.
For this work, we use custom built robots (as in \cite{pulkit}) with a linear actuator (piston) mounted on their chassis.
Our approach involves an open loop control for manipulation the piston height to minimize the error between the desired and the current orientation of the payload.
In this work, we aim to keep the payload level with the ground (\ang{0} roll and pitch).
However our proposed solution can be used for any desired orientation subject to system constraints.
\cite{pulkit} has shown formation control on smooth surface and our work will allow us multi-robot system to carry the payload on it top maneuver over irregular surface in the formation while maintaining the payload level horizontal to ground.

\section{MODELING AND CONTROL}
This section deals with a discussion on kinematics model, piston design and an approach for making multiple robots to move in a loosely coupled formation.
\subsection{Kinematics Model}
Each agent (robot) is a four-wheeled non-holonomic robot.
We assume a skid steer kinematic model which is systematically described in \cite{skidsteer}.
In essence, we assume both the wheels on each side move at the same speed.
Each agent has a generalized state $ q = [x, y, \theta]^T, $ where $(x, y)$ is the position of the robot in the global inertial frame, and $\theta$ is the yaw of the robot. Control input to each robot is given as $ u = [v, \omega]^T$, where $v$ is the linear velocity and $\omega$ is the angular velocity of the robot.
The evolution of the state of the robot with time is given by the following equation:
\begin{equation}
    q(t+1) = A*q(t) + B(t)*u(t)
\end{equation}
\begin{center}
    where, $A = \begin{bmatrix} 1 & 0 & 0\\
                            0 & 1 & 0\\
                            0 & 0 & 1 
                \end{bmatrix}
            $
\end{center}
\begin{center}
          and $B = \begin{bmatrix} \cos(\theta(t))*dt & 0\\
                            \sin(\theta(t))*dt & 0\\
                            0 & dt
            \end{bmatrix}
            $
\end{center}

\begin{figure}[h]
    \begin{center}
        \includegraphics[scale=0.32]{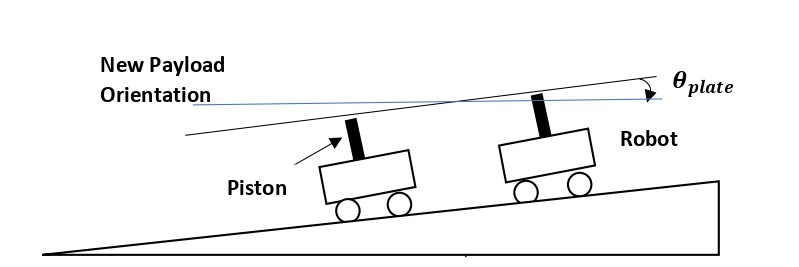}
    \caption{\small Robots with Linear Actuator carrying a payload on inclined plane where the black line shows orientation when both robots have same piston length while the blue line shows desired orientation of the payload.}
    \label{model1}
    \end{center}
\end{figure}

\subsection{Piston Model}
\begin{figure}[h]
\begin{center}
    \includegraphics[scale=0.18]{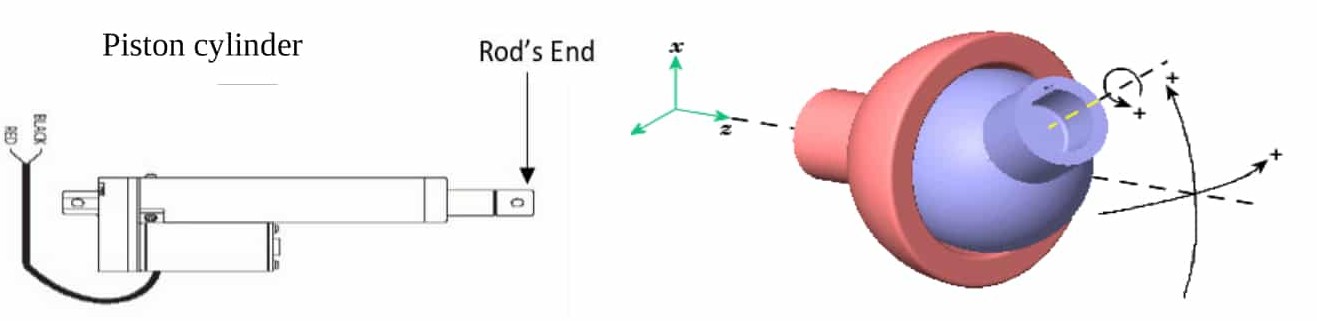}
    \newline
    \caption{\small Piston model showing rod's end and spherical ball joint which we propose as an interface between the piston and the piston plate.}
    \label{pis}
\end{center}
\end{figure}

A linear actuator is mounted on top of the robot chassis.
This actuator has four major parts: a hollow cylinder, piston rod, spherical ball joint and a rectangular plate.
The orientation of payload can be manipulated by moving the pistons of the robots up and down.
At the end of the piston's rod, a small rectangular plate is attached via a spherical ball joint.
The advantage of the spherical ball joint is that it allows free rotation about all three axis while preventing any translation motion.
The spherical ball joint will help the rectangular plate to self-adjust the angle between the payload and the piston's rod.
The schematic of the piston and joint is showed in Figure \ref{pis}.
The rectangular plate increases the surface contact to the payload, thus avoiding the slippage of payload as shown in Figure \ref{pisangle}.

\subsection{Formation Control}
To carry the payload as a multi-robot team, we employ a bio-inspired formation control as described in \cite{pulkit}.
The formation control is based on a leader follower paradigm.
The leader of the team is assigned a predefined trajectory, which is assumed to be generated by a global motion planner \cite{OMPL}.
The follower robots derive their respective command velocities from the leader’s position and command velocities by using the control law defined by the following equations.
\begin{align*}
    v_j &= k_1\alpha_j + v_icos\theta_{ij} - \rho_{ij}^d\omega_isin(\psi_{ij}^d - \theta_{ij})\\
    \omega_j &= (v_isin\theta_{ij} + \rho_{ij}^d\omega_icos(\psi_{ij}^d + \theta_{ij}) + k_2\beta_j + k_3\theta_{je}) / d
\end{align*}
where $v_j$ and $\omega_j$ are the linear and angular velocities of the $j^th$ robot, $\alpha_j$ and $\beta_j$ is the error along $x$ and $y$ axis respectively, $k_1, k_2, k_3, > 0$ are the control gains, $\rho^d_{ij}$ and $\psi_{ij}$ is the desired distance and orientation to maintain between robot $i$ and robot $j$, $x_{je}$, $y_{je}$ and $\theta_{je}$ are the positional tracking errors between the leader and follower.
\section{PROBLEM FORMULATION}
We assume that the orientation of the payload is known to us at every time step.
This can be achieved by attaching a small device with an Inertial Measurement Unit (IMU) onto the payload.
To keep the payload horizontal with respect to the ground, we control the height of the piston.
Mathematically, we wish to keep $\theta_{roll}$ and $\theta_{pitch}$ of payload as close to zero as possible.\\
\par
The IMU sensor mounted on the payload will measure the orientation of payload and our model will use this orientation value to find amount by which the piston of each robot should adjust its height in order to keep the payload level horizontal.
For every possible orientation of the payload in roll-pitch space, we have a unique solution of piston's height.
If we consider a plane along the payload's  $x-y$ plane, then the solution  of the pistons height would be the $z$ value of points lying on that plane having the $x$, $y$ coordinates of the piston.
The roll and pitch angle measured, is feed with negative value because when the payload passes through an inclined path, then it should be tilted by the same angle but in opposite direction.
\par
In 1-D if a line is rotated about its origin O, the arc length generated at given distance from origin is given by:
\begin{equation*}
    \Delta l = \theta * r
\end{equation*}
where, $\Delta l$ is the arc length, $r$ is the distance  from origin, $\theta$ is the angle (in radians) of payload. Assuming $\theta$ is very small, the arc length can be taken as a straight line.

\par
Let us consider a 2-D Cartesian coordinate system for the payload, where $(x, y)$ represent robot's position in that system. 


If the payload is rotated by a small angle only about the x-axis, then the arc length subtended at this robot's location can be given as:
\begin{equation}
    \vec{\Delta l_x} = \theta_x * y
\end{equation}

Similarly, if the robot is rotated only about the y-axis, then the arc length subtended can be given as:
\begin{equation}
    \vec{\Delta l_y} = \theta_y * x
\end{equation}

As both these vectors are along the z-axis, their vector sum will be equal to their scalar sum.
Combining the two, the equivalent arc length can be given as:
\begin{equation}
    \Delta l = \vec{\Delta l_x} + \vec{\Delta l_y} = (\theta_x * y) + (\theta_y * x)
\end{equation}
where, $\Delta l$ can be either positive or negative.
To make sure that the payload is able to roll and pitch in both anti-clockwise and clockwise directions, we define a mean piston height which is the half the maximum achievable height of the piston.
If total height that can be attained by piston is represented by $l_{total}$, then mean height of the piston rod can be given as $l_{mean} = l_{total}/2$.
This $l_{mean}$ will be the initial height of piston rod when robots will move on horizontal surface, thus enabling the payload to be rotated in any possible direction up to the limits of the piston.
The maximum and minimum difference in angle that can be achieved by this multi-robot team to adjust the payload's orientation is described below.
\par
As the piston length has a maximum bound $l$ we can manipulate the payload only up to a certain extent.
Consider a formation of $N$ robots, with robots at positions described by $(x_i, y_i)$, then the maximum/minimum roll $(\theta_x^{max}, \theta_x^{min})$ and pitch $(\theta_y^{max}, \theta_y^{min})$ angles that can be achieved on a smooth surface are defined below:\\
\begin{center}
$\frac{-l}{max(x_i) - min(x_i)} \leq \theta_{y} \leq \frac{l}{max(x_i) - min(x_i)}$\\
$\frac{-l}{max(y_i) - min(y_i)} \leq \theta_{x} \leq \frac{l}{max(y_i) - min(y_i)}$
\end{center}


For $n$ robots placed at $(x_i, y_i), \forall i \in 0, 1, ... n$, the change in piston length required to keep the payload level with the ground, given the current payload orientation $\theta$ can be written as:
\begin{equation}
    \Delta L = B * \theta
\end{equation}
where $B$ and $\theta$ are given by
\begin{equation*}
B = \begin{bmatrix}
y_0 & x_0\\
y_1 & x_1\\
\vdots & \vdots\\
y_{n-1} & x_{n-1}
\end{bmatrix}\\
\theta = \begin{bmatrix}
\theta_x\\
\theta_y
\end{bmatrix}
\end{equation*}
and the $i^th$ value in the vector $\Delta L$ represents the amount of change required in the height of the piston of $i^th$ robot.
The proposed algorithm that is we employ to maintain the payload orientation is given below 
Let $L$ be the vector of the desired piston heights of all the robots, $L_{mean}$ be the vector of all mean piston lengths, then our algorithm is shown below.

\begin{algorithm}[]
\caption{Finding Piston's Rod Height}
\label{alg:algo1}
\SetAlgoLined
\textbf{Initialisation}\\
$L_{previous}  \gets  L_{mean} $\\
$L \gets L_{previous}$; \Comment{Initial Piston's Rod Height}\\
\While {true}{
    read($ \theta$); \Comment{get current roll \& pitch angle} \\
    $\Delta L \gets B * \theta$;\\
    $L \gets L_{previous} + \Delta L$;\\
    $L_{previous} \gets L$;\\
}
\end{algorithm}


\section{METHODOLOGY}
\begin{figure}[h]
\begin{center}
    \includegraphics[scale=0.20]{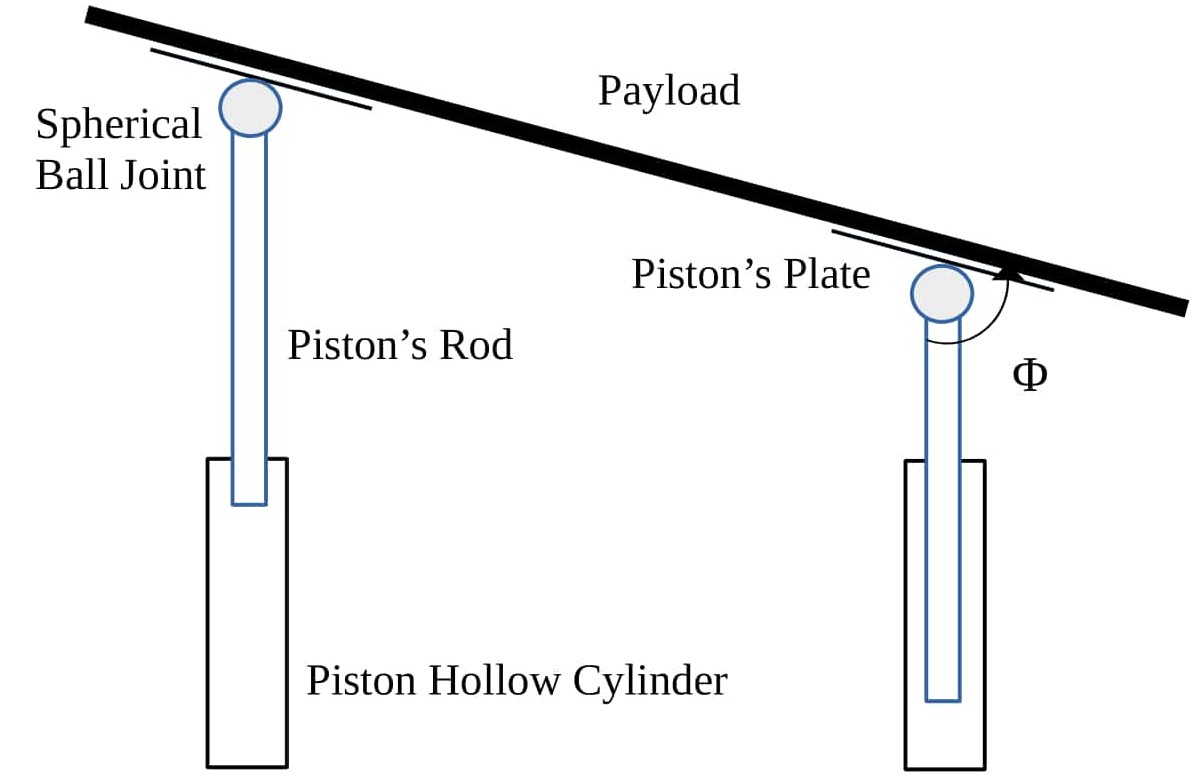}
    \caption{\small Angle $\phi$ created between piston rod and the plate. $\phi$ can be manipulated by changing the rod length of different robots.}
    \label{pisangle}
    \end{center}
\end{figure}
\begin{figure}
\begin{center}
    \includegraphics[scale=0.5]{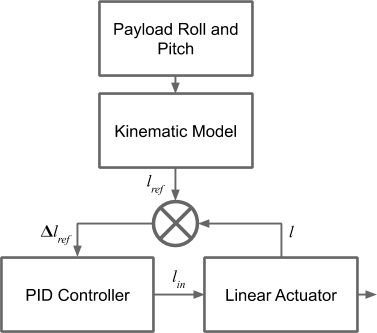}
    \caption{Block diagram representing the open loop control system that utilises the roll and the pitch angle of the payload to achieve required piston lengths through PID controller. $l_{ref}$ is reference piston height, $\Delta l_{ref}$ is difference in current and desired piston length and $l_{in}$ is input length feed to piston.}
    \label{flowchart}
\end{center}
\end{figure}

Our proposed solution assumes that the orientation of the payload is know.
This could be achieved via various methods like pose estimation using a motion capture system or by an IMU placed on the payload.
All the robots along with the pose estimation node are assumed to be able to communicate with each other. 
Because the change in a robot's piston's height is independent of the other robot's state, therefore each robot can asynchronously compute their desired controls.
Each robot is provided with its relative position with respect to the payload and can calculate the desired control based on its row vector in $B$.
\par
To control the orientation of the payload, each of the robot has to move its piston accordingly.
The model proposed above gives the desired height of the piston in order to level the payload with the $xy$ plane.
A PID \cite{c5} controller is used to achieve this height.
\begin{equation}
u(t) = K_pe(t) + K_I\int_{0}^{t}e(\tau)d\tau + K_D \frac{de(t)}{dt}
\end{equation}
where:\\
$K_p$ : proportional gain\\
$K_I$ : integral gain\\
$K_D$ : derivative gain\\
$e(t)$ : difference between desire position and current position of piston's rod length\\

The piston controllers use a closed loop feedback mechanism, in which it will continuously find difference between current piston's length and required piston's length, after that applies a correction based on proportional, integral, and derivative terms so that the difference becomes zero. Based on the difference, either positive or negative, the piston will automatically move up or down.
\par
We use Gazebo simulator to run experiments on our proposed solution.
For inter-robot communications we use ROS (Robotic Operating System).
We also used an IMU plugin for Gazebo which could broadcast the orientation of the payload to all the ROS nodes.
The frequency of the IMU was set to $50$Hz which is close the actual hardware ones.
\par
A crucial part of the problem is to control the orientation of the payload for all orientations of the robot.
A simple piston design, such as in \cite{pulkit}, would not be able to make full contact with the payload if the robots were inclined.
To overcome this issue we propose a ball and socket attachment to the piston.
This helps the payload maintain constant contact with the plate at the end of the piston.
Figure \ref{pisangle} shows a situation when the angle between piston's plate and rod $\phi$ will be automatically adjusted due to a free spherical ball joint  without employing any external actuator. Without loss of generality, $\phi$ can be defined for both roll and pitch of the payload.
To the best of our knowledge, there was no such joint in the Gazebo plugin library, so to achieve this we employed three orthogonal revolute joints.

\section{RESULTS}

\begin{figure}
	\centering
    \includegraphics[scale = 0.45]{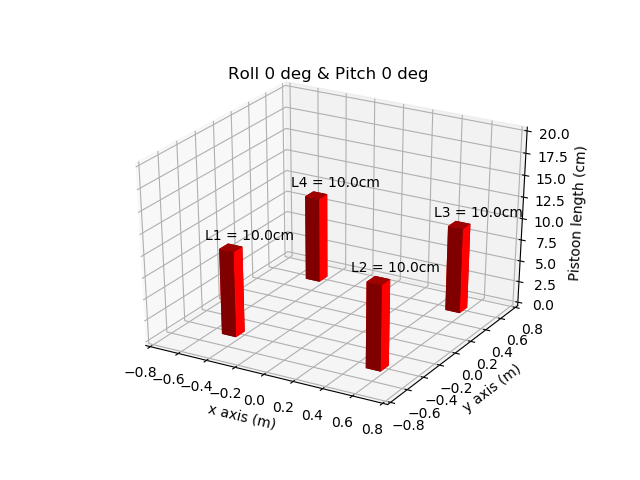} 
    \caption{\small Height of piston at Roll = $\ang{0}$ and Pitch = $\ang{0}$.} \normalsize
    \label{res1}
\end{figure}
\begin{figure}
	\centering
    \includegraphics[scale = 0.45]{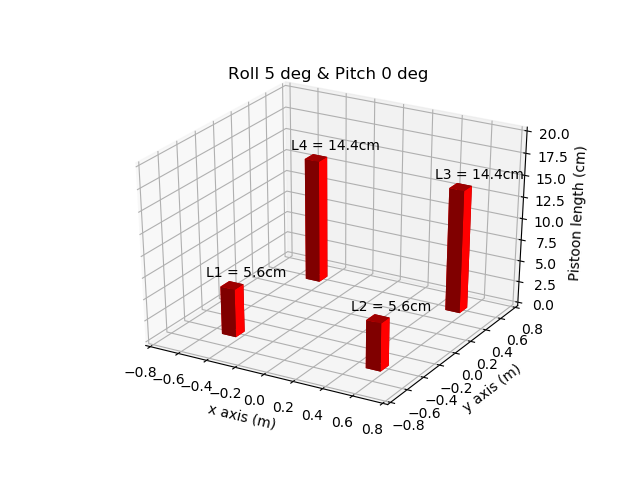} 
    \caption{\small Height of piston at Roll = $\ang{5}$ and Pitch = $\ang{0}$.} \normalsize
    \label{res2}
\end{figure}
\begin{figure}[h]
	\centering
    \includegraphics[scale = 0.45]{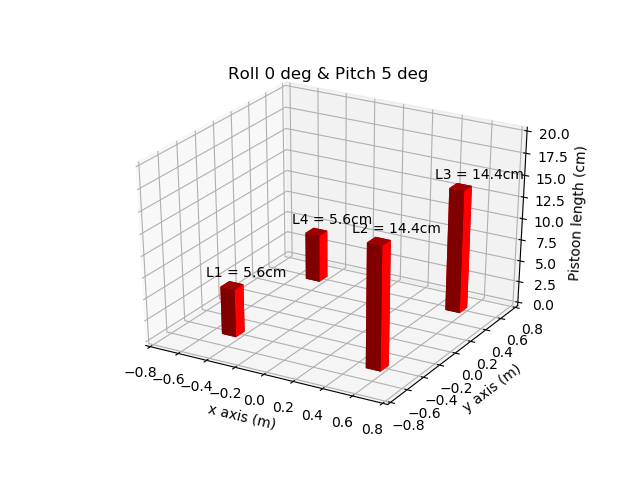} 
    \caption{\small Height of piston at Roll = $\ang{0}$ and Pitch = $\ang{5}$.} \normalsize
    \label{res3}
\end{figure}
\begin{figure}[h]
	\centering
    \includegraphics[scale = 0.45]{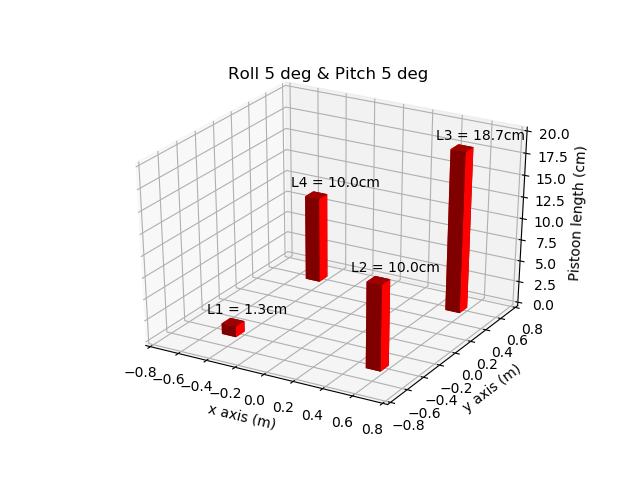} 
    \caption{\small Height of piston at Roll = $\ang{5}$ and Pitch = $\ang{5}$.} \normalsize
    \label{res4}
\end{figure}

\begin{figure}
	\centering
    \includegraphics[scale = 0.22]{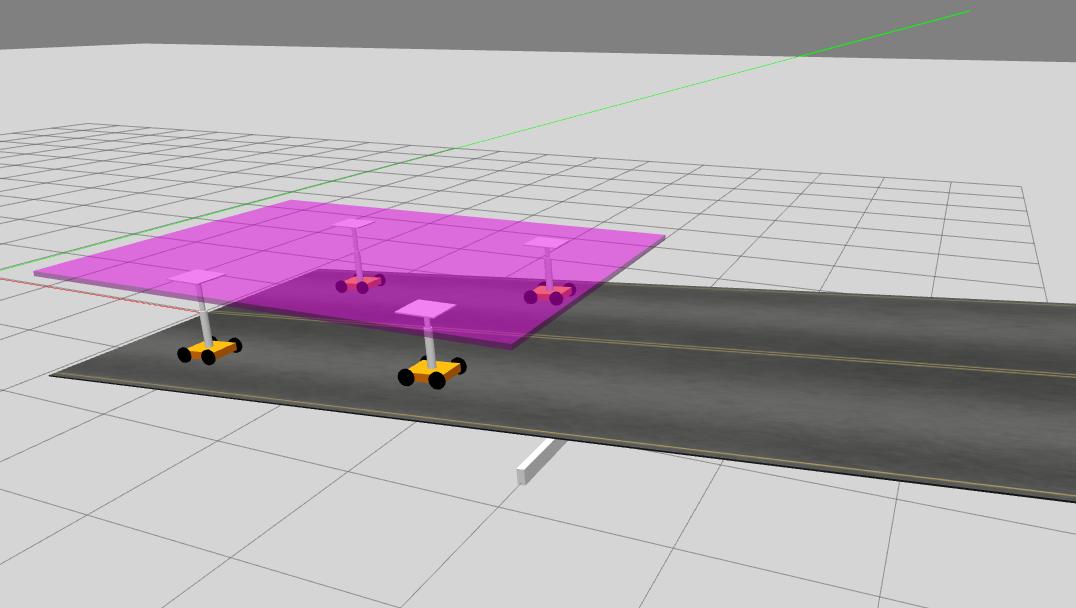} 
    \caption{\small Snapshot of Gazebo simulation of $4$ robots carrying payload on an inclined surface.} \normalsize
    \label{fig-1}
\end{figure}
\begin{figure}
	\centering
    \includegraphics[scale = 0.22]{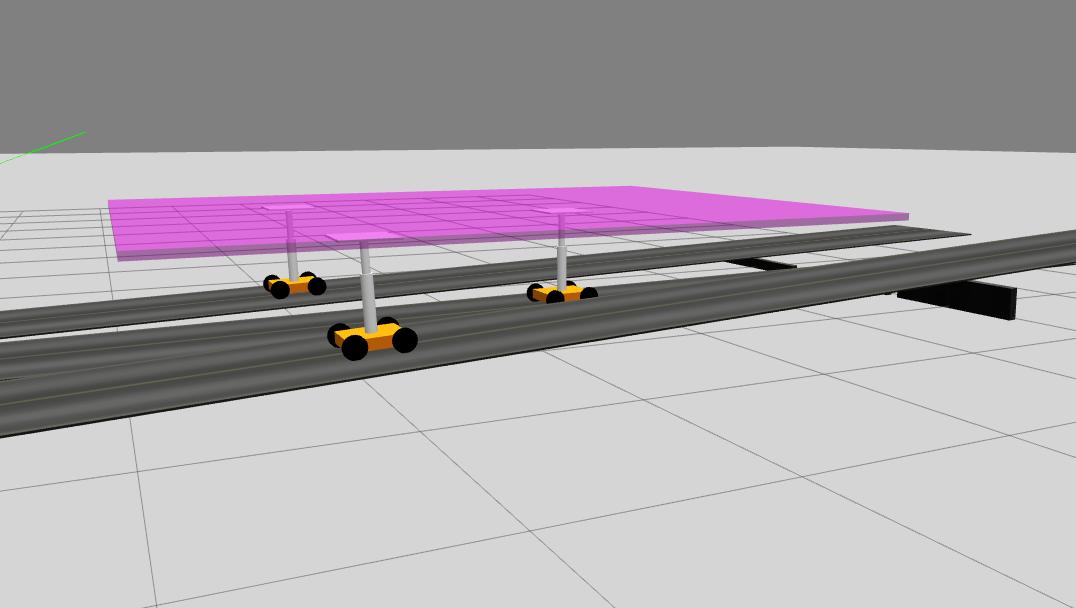} 
    \caption{\small Snapshot of three robots moving on three different inclined surfaces.} \normalsize
    \label{fig-2}
\end{figure}
\begin{figure}[h]
	\centering
    \includegraphics[scale = 0.22]{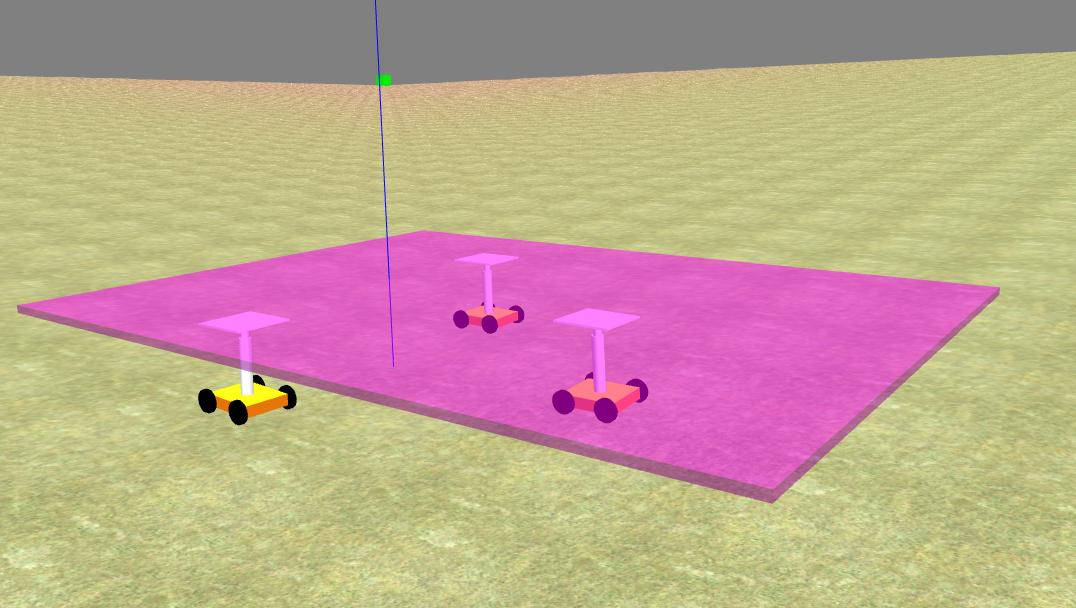} 
    \caption{\small Snapshot of three robots carrying the payload in Gazebo's \emph{heightmap.world} having uneven terrain.} \normalsize
    \label{fig-3}
\end{figure}
\begin{figure}[h]
	\centering
    \includegraphics[scale = 0.22]{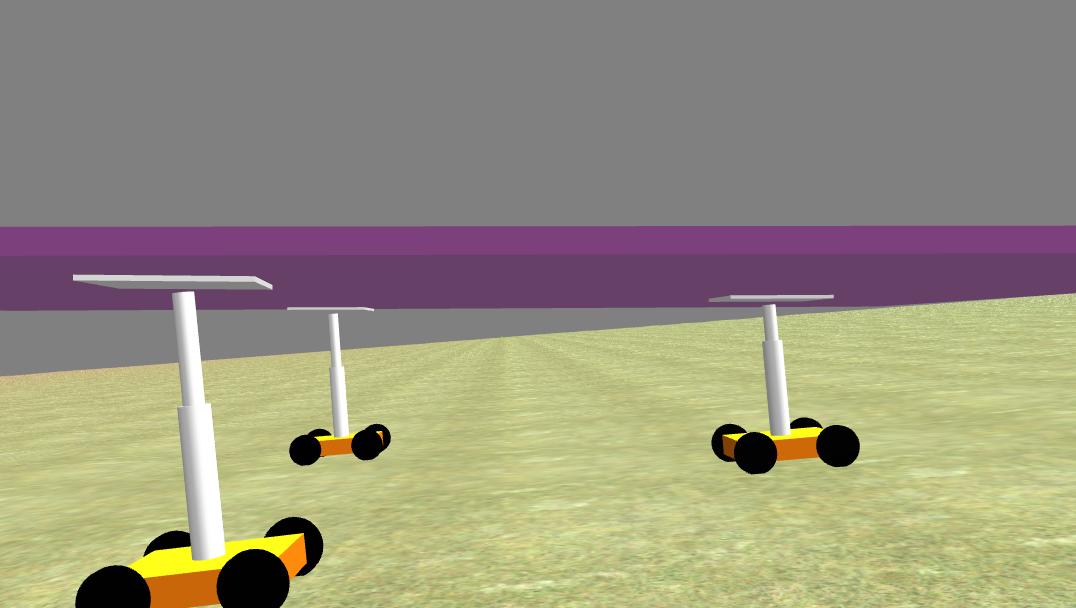} 
    \caption{\small Self-adjustment of the angle between piston plate and piston rod due to spherical ball joint.} \normalsize
    \label{fig-4}
\end{figure}



\begin{table}
    \begin{center}
        \begin{tabular}{ |c|c| } 
         \hline
         Payload Mass ($M_{payload}$) & $5Kg$ \\ 
         Mean Piston Length ($l_{mean}$) & $0.125m$ \\ 
         Robot Mass ($M_{robot}$) & $5Kg$ \\
         Robot Dimensions ($a$x$b$) & $30cm$ $\times$ $20cm$\\
         \hline
        \end{tabular}
        \caption{\small Payload and Robot parameters used in our simulations.}
        \label{table:1}
    \end{center}
\end{table}

We simulated the payload balancing task for multiple robot formations. We have used Python for implementing the proposed control solution and Gazebo for simulation to verify our results.
We defined, $\theta_x$ as roll angle and $\theta_y$ as pitch angle.
We use our custom designed four wheeled robots \footnote{https://github.com/aarg-kcis/minion-ros-gazeobo-rviz/}.
All robots have a piston which has been modelled as a prismatic joint in Gazebo.
At one end of this piston, a rectangular plate has been attached using a ball and socket joint.
Figure \ref{fig-4} shows the setup of a three robot team carrying a payload. Note that the rectangular plate is not aligned with the ground.
\par
\begin{figure*}%
\centering
\begin{subfigure}{\columnwidth}
\includegraphics[width=\columnwidth]{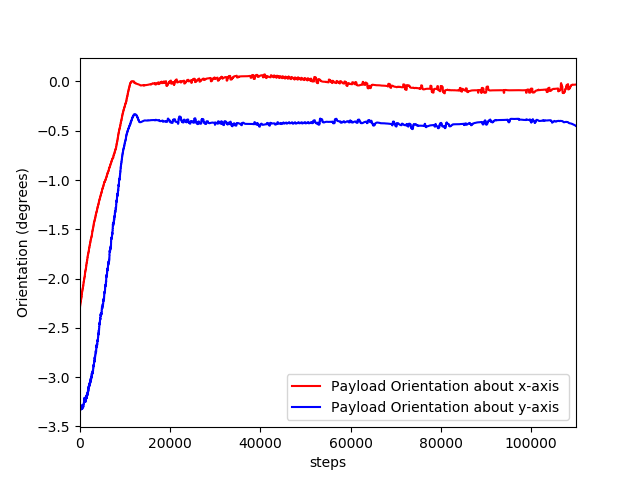}%
\caption{}%
\label{psubfiga}%
\end{subfigure}\hfill%
\begin{subfigure}{\columnwidth}
\includegraphics[width=\columnwidth]{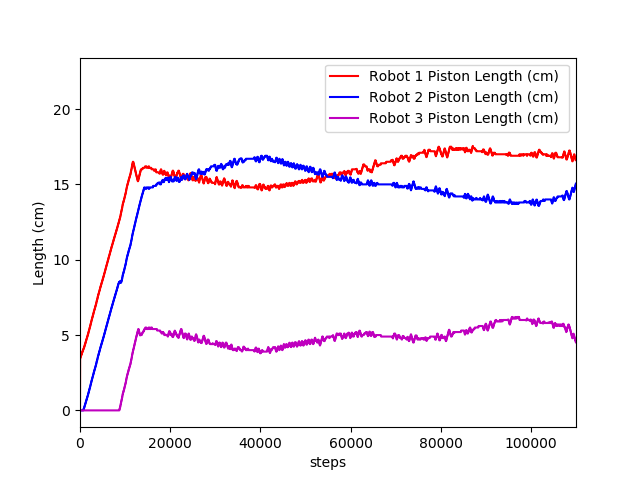}%
\caption{}%
\label{psubfigb}%
\end{subfigure}\hfill%
\caption{(a) Roll and pitch of the payload at different time steps. (b) Height of the pistons of each robot while carrying the payload in the \emph{heightmap.world} map.}
\label{roboto}
\end{figure*}

We tested our algorithm in three different situations as described below.
\begin{enumerate}
    \item Four robots were used to carry a rectangular payload on a plank of fixed inclination (pitch only). Figure \ref{fig-1}
    \item Three robots move in a triangular formation to carry a payload. All the robots move on three different planks with different but fixed slopes. Figure \ref{fig-2}
    \item Three robots move in a triangular formation to carry a payload. In this case, \emph{heightmap.world} world of gazebo is used having  uneven surface. Figure \ref{fig-3}
\end{enumerate}
All robots go through different roll and pitch angles when they traverse the arena.
The robots were placed at locations (-0.5,1), (-0.5,-1), (0.5,0.5) with respect to the payload's reference frame.
Figure \ref{psubfiga} shows the orientation of the payload with respect to time.
We can see roll angle (red) and pitch angle (blue), both are almost zero all the time steps. Initially robots are at rough surface, so the payload placed on top of robot has some non-zero roll and pitch value. Once algorithm start running roll and pitch angle tends to zero.

Figure \ref{psubfigb} shows different heights attained by the pistons of robot 1, robot 2 and robot 3 with respect to time.
Initially all piston rods are at zero height.
This is intentional and the robots move their pistons up to the mean height before they can start balancing the payload. This can be seen in the plots as the heights of the pistons start increasing and attains the mean height.
\par
In figure \ref{roboto}, the red line and the blue line denote roll and pitch respectively for robot 1, robot 2 and robot 3 with respect to time.
The continuous change in angle shows that surface in the \emph{heightmap.world} is rough and irregular, so orientation of robot changes with the time.



\begin{figure*}
\centering
\begin{subfigure}{0.68\columnwidth}
\includegraphics[width=\columnwidth]{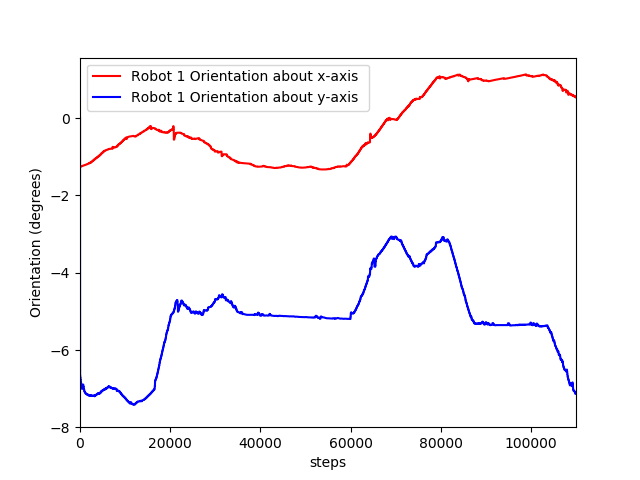}%
\caption{Robot 1}%
\label{rrsubfiga}%
\end{subfigure}\hfill%
\begin{subfigure}{0.68\columnwidth}
\includegraphics[width=\columnwidth]{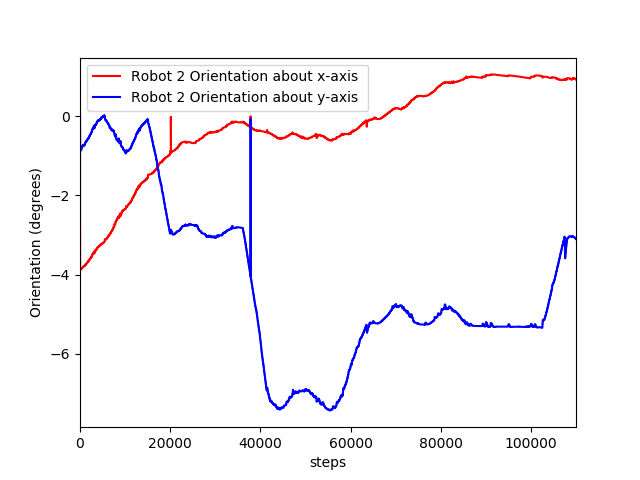}%
\caption{Robot 2}%
\label{rrsubfigb}%
\end{subfigure}\hfill%
\begin{subfigure}{0.68\columnwidth}
\includegraphics[width=\columnwidth]{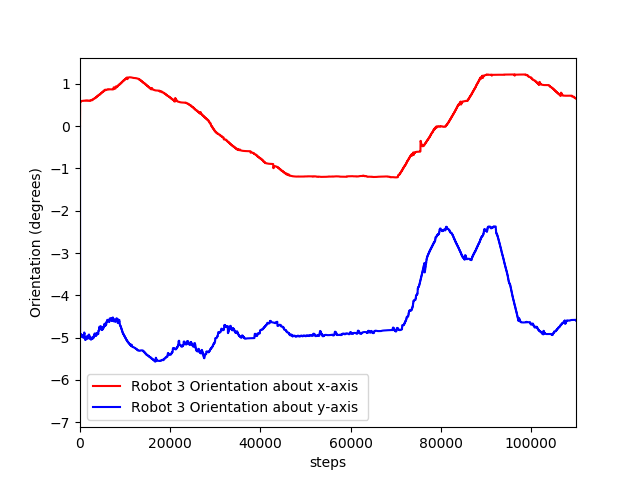}%
\caption{Robot 3}%
\label{rrsubfigc}%
\end{subfigure}
\caption{Roll and pitch of different robots at different time steps while carrying the payload in the \emph{heightmap.world} map.}
\label{roboto}
\end{figure*}


\section{CONCLUSIONS AND FUTURE WORK}
Through this work, we have shown the feasibility of our proposed solution for multi-robot system carrying a payload in formation on irregular surfaces while keeping the payload orientation level with the ground. We have used a loosely coupled multi-robot system to accomplish the task. Each robot in our multi-agent system is non-holonomic and working independently to maintain the formation and orientation of the payload.
In addition to this, the scope of future work would lie in making this system more dynamic and robust.
One such way of achieving this would be to place a low cost stereo camera or a depth sensing camera like Kinect to scan the surface of the terrain ahead and make decisions based on the surface ahead.

\bibliographystyle{ieeetr}
\bibliography{root}
\end{document}